%% file: main.tex
\documentclass{article}

\usepackage{microtype}
\usepackage{graphicx}
\usepackage{subfigure}
\usepackage{booktabs} % for professional tables

\usepackage{hyperref}

\usepackage[accepted]{icml2025_preprint}

\usepackage{amsmath}
\usepackage{amssymb}
\usepackage{mathtools}
\usepackage{amsthm}
\usepackage{pifont}

\usepackage[capitalize,noabbrev]{cleveref}

\theoremstyle{plain}

\theoremstyle{definition}

\theoremstyle{remark}

\usepackage[textsize=tiny]{todonotes}

\icmltitlerunning{A Simple and Generalist Approach for Panoptic Segmentation}

\begin{document}

\twocolumn[
% \icmltitle{Submission and Formatting Instructions for \\
%            International Conference on Machine Learning (ICML 2025)}
\icmltitle{A Simple and Generalist Approach for Panoptic Segmentation}



\icmlsetsymbol{equal}{*}

\begin{icmlauthorlist}
\icmlauthor{Nedyalko Prisadnikov}{insait}
\icmlauthor{Wouter Van Gansbeke}{equal}
\icmlauthor{Danda Pani Paudel}{insait}
\icmlauthor{Luc Van Gool}{insait}
\end{icmlauthorlist}

\icmlaffiliation{insait}{INSAIT, Sofia University "St. Kliment Ohridski"}
%\icmlaffiliation{todo}{Supervised the first author while at INSAIT, now affiliated with Google DeepMind}
% \icmlaffiliation{eth}{ETH Zurich}

\icmlcorrespondingauthor{Nedyalko Prisadnikov}{first.last@insait.ai}

\icmlkeywords{Machine Learning, ICML}

\vskip 0.3in
]



\printAffiliationsAndNotice{\icmlEqualContribution} % otherwise use the standard text.

\input{sec/0_abstract}
\input{sec/1_introduction}
\input{sec/2_related_work}
\input{sec/3_method}

\input{sec/4_rationale}
\input{sec/5_experiments}
\input{sec/6_conclusion}

% % In the unusual situation where you want a paper to appear in the
% % references without citing it in the main text, use \nocite

\bibliography{main}
\bibliographystyle{icml2025}

%%%%%%%%%%%%%%%%%%%%%%%%%%%%%%%%%%%%%%%%%%%%%%%%%%%%%%%%%%%%%%%%%%%%%%%%%%%%%%%
%%%%%%%%%%%%%%%%%%%%%%%%%%%%%%%%%%%%%%%%%%%%%%%%%%%%%%%%%%%%%%%%%%%%%%%%%%%%%%%
% APPENDIX
%%%%%%%%%%%%%%%%%%%%%%%%%%%%%%%%%%%%%%%%%%%%%%%%%%%%%%%%%%%%%%%%%%%%%%%%%%%%%%%
%%%%%%%%%%%%%%%%%%%%%%%%%%%%%%%%%%%%%%%%%%%%%%%%%%%%%%%%%%%%%%%%%%%%%%%%%%%%%%%
\newpage
\appendix
\onecolumn
\input{sec/a1_depth}
\input{sec/a2_losses}
\input{sec/a3_encodings}
\input{sec/a4_qualitative}
% You can have as much text here as you want. The main body must be at most $8$ pages long.
% For the final version, one more page can be added.
% If you want, you can use an appendix like this one.  
% 
% The $\mathtt{\backslash onecolumn}$ command above can be kept in place if you prefer a one-column appendix, or can be removed if you prefer a two-column appendix.  Apart from this possible change, the style (font size, spacing, margins, page numbering, etc.) should be kept the same as the main body.
%%%%%%%%%%%%%%%%%%%%%%%%%%%%%%%%%%%%%%%%%%%%%%%%%%%%%%%%%%%%%%%%%%%%%%%%%%%%%%%
%%%%%%%%%%%%%%%%%%%%%%%%%%%%%%%%%%%%%%%%%%%%%%%%%%%%%%%%%%%%%%%%%%%%%%%%%%%%%%%

\end{document}

%% file: sec/0_abstract.tex
\begin{abstract}
Panoptic segmentation is an important computer vision task,
where the current \textit{state-of-the-art} solutions
require specialized components to perform well.
We propose a simple generalist framework
based on a deep encoder - shallow decoder architecture
with per-pixel prediction.
Essentially fine-tuning a massively pretrained image model with minimal additional components.
Naively this method does not yield good results.
We show that this is due to imbalance during training
and propose a novel method for reducing it -
centroid regression \emph{in the space of spectral positional embeddings}.
Our method achieves panoptic quality (PQ) of \textbf{55.1} on the challenging MS-COCO dataset,
\textit{state-of-the-art} performance among generalist methods.
\end{abstract}

%% file: sec/1_introduction.tex
\section{Introduction}

Panoptic segmentation~\cite{kirillov2019panoptic} is a dense prediction computer vision task
that combines semantic and instance segmentation.
It consists of labeling each pixel of an input image with its corresponding semantic category,
as well as the specific instance to which it belongs.
It tackles both \textit{stuff-like} instance-less categories (like sky, grass, etc)
and \textit{thing-like} categories with potentially multiple instances per image (people, cars, etc).
Solving segmentation tasks is useful for numerous real world applications including
autonomous driving~\cite{cordts2016cityscapes,mohan2021efficientps},
agriculture~\cite{darbyshire2023hierarchical,chiu2020agriculture},
medical imaging~\cite{zhang2018panoptic,menze2014multimodal},
etc.

The instance segmentation part of the task requires solutions to be permutation invariant.
Swapping the labels between two instances of the same class is still a valid solution.
This poses a significant challenge.
Earlier solutions required specialized components like region proposal networks~
\cite{he2017mask,vasconcelos2017cascade}.
Recent \textit{state-of-the-art} approaches~\cite{jain2023oneformer,maskdino,cheng2021per,yu2022kmax}
propose an end-to-end strategy by casting the problem as a direct set prediction task.
They achieve strong results but require a complex loss function based on bipartite matching,
and contain transformer decoders with object queries~\cite{carion2020end},
specialized for segmentation and detection tasks.

The vision community vouch for the use of generalist architectures~
\cite{lu2022unified,lu2024unified,wang2023images}
that can lead to models generalizing across multiple tasks.
% Recent generalist models
% like Unified-IO~\cite{lu2022unified,lu2024unified} and Painter~\cite{wang2023images}
% vouch for using task-agnostic architectures and loss functions.
% This can lead to simple models that generalize across multiple tasks.
We propose a very simple deep encoder - shallow decoder framework
with per-pixel prediction for the panoptic segmentation task.
Using a strong pretrained image model (DINOv2~\cite{dinov2}) as an encoder,
and adding minimal architectural components to make it suitable for dense tasks.
Namely,
a shallow decoder that upscales the patch level representations to pixel level ones.
Using a per-pixel prediction with averaged pixel loss
makes it easy to apply this framework to different dense tasks.

This approach has worked well for some problems,
but lags behind for panoptic segmentation.
We will show that naively applying this framework for panoptic segmentation
suffers from inherent imbalance during training.
Understanding the source of it,
we propose simple solutions that enable us
to bridge the gap between generalist and specialized methods.
% We propose simple solutions that make this approach produce results close
% to the \textit{state-of-the-art}.
Our main \underline{contributions} are the following.

\begin{itemize}
    \item We bridge the gap between specialized and generalist models using a simple
    method that can be described as fine-tuning DINO, while also performing well in a multi-tasking setup.
    \item We propose a novel approach (Section~\ref{sec:pos_embed})
    for reducing the imbalance
    between small and large objects
    while training instance segmentation through centroid regression.
    % in the spectral positional embeddings space.
\end{itemize}

With per-pixel prediction,
the permutation invariance of the instance labels
can be addressed with deterministic labeling.
We follow~\cite{kendall2018multi,cheng2020panoptic,wang2023images}
and cast the instance segmentation problem as a pixel regression one.
For each pixel we predict the location of the center of mass
of the instance it belongs to.

We show that performing direct centroid regression on the pixel level
is prone to significant imbalance while training.
By making a prediction of the spectral sine-cosine positional embeddings of the centroid coordinates instead
(Section~\ref{sec:pos_embed}),
we lift them to a higher-dimensional space where the imbalance is reduced.
We give a detailed rationale behind this approach (Section~\ref{sec:motivation}),
backed by strong experimental evidence (Section~\ref{sec:experiments}).

Additionally,
we use a boundary aware loss modulation~\cite{zhen2019learning,zhang2019deep,vsaric2023panoptic,zhu2019improving}
in the form of edge distance sampling (EDS).
This helps us to focus on \textit{harder} pixels.
We further show that EDS helps minimize
the training imbalance between small and large objects.

With these we set a new \textit{state-of-the-art} performance
for panoptic segmentation among generalist methods.
We achieve panoptic quality of \textbf{55.1} on MS-COCO~\cite{lin2014microsoft}.

% With this we set a new \textit{state-of-the-art} result for panoptic segmentation
% on the MS-COCO dataset,
% with panoptic quality of \textbf{55.1 PQ},
% while also performing strongly in a multitasking setup with panoptic segmentation and depth prediction.

%% file: sec/2_related_work.tex
\section{Related work}

\paragraph{Panoptic segmentation}
is one of the more challenging segmentation tasks.
Semantic segmentation can be solved by
simply classifying each pixel independently.
With instance segmentation such a simple approach
is not directly applicable.
On top of classifying each pixel,
we need to cluster pixels belonging to the same instance in a permutation invariant way.
This challenge has forced solutions to use custom components.
\textit{State-of-the-art} methods are based on DETR~\cite{carion2020end}
- a transformer decoder with object queries.
They cast the problem as direct set prediction.
The model outputs a set of masks,
then a bipartite matching is applied
between the predicted and target masks
using Hungarian matching~\cite{kuhn1955hungarian}.
% The model outputs a set of masks.
% Then they perform bipartite matching between the predicted and ground truth masks
% using Hungarian matching~\cite{kuhn1955hungarian}.
This works very well,
but requires a complicated loss mechanism.
Also, the majority of the model's parameters and computational complexity
lie in the transformer decoder,
which is specialized for generating segmentation masks.
With the appearance of strong ViT-based~\cite{vit} image representation models,
we ask ourselves if it is possible to reuse these strong general representations
with minimal additional architectural components.

% This resolves the permutation invariance of the instance labels.
% Despite using a complicated loss mechanism,
% the majority of the model parameters and computation complexity
% is in the transformer decoders specialized in prediction the segmentation masks.
% With the advent of powerful image representation models like DINO~\cite{dinov2},
% we ask ourselves if it is possible to reuse the strong general representations from them
% with minimal specialized architectural components.

\paragraph{Denoising diffusion methods.}
Recently~\cite{chen2023generalist} and~\cite{van2024simple}
proposed using denoising diffusion for generating panoptic masks,
as a generalist framework.
During training,
they generate the target masks using random instance IDs with noise on top of them.
The model learns to predict that specific random assignment of the label IDs.
During inference the model is given random input noise
and generates random assignment of the instance labels.
This prevents the necessity of bipartite matching in the training loop.
However, it requires multiple denoising iterations to perform well.

\paragraph{Methods with pixel-wise prediction.}
If we want to use per-pixel prediction for panoptic or instance segmentation,
we need a deterministic label for each instance.
That way for each pixel we have a stable label to predict.
A popular approach is using the coordinates of the center of mass of the instance as a label proxy~
\cite{kendall2018multi,cheng2020panoptic,wang2023images}.
We base our work on this idea.
Centroid regression has been used for panoptic segmentation in~\cite{wang2023images},
but the gap to the \textit{state-of-the-art} methods is significant.
We will show simple modifications that unlock strong performance.
% We will show that directly regressing the centroid coordinates leads to imbalance and challenges during training,
% and propose simple solutions for them.

\paragraph{Boundary aware per-pixel loss modulation}
is a method used with pixel-wise loss to focus on the more challenging parts of the images~
\cite{zhen2019learning,zhang2019deep,vsaric2023panoptic,zhu2019improving},
i.e. the pixels near object boundaries.
We show that this approach also diminishes an imbalance between small and large objects during training.
We perform Edge Distance Sampling (EDS) which only samples pixels near instance boundaries
when computing the loss.
The weight of each pixel is determined based on the distance to the nearest boundary.

%% file: sec/3_method.tex
\section{Method}

% \begin{figure}[ht]
% \vskip 0.2in
% \begin{center}
% \centerline{\includegraphics[width=\columnwidth]{icml_numpapers}}
% \caption{Historical locations and number of accepted papers for International
% Machine Learning Conferences (ICML 1993 -- ICML 2008) and International
% Workshops on Machine Learning (ML 1988 -- ML 1992). At the time this figure was
% produced, the number of accepted papers for ICML 2008 was unknown and instead
% estimated.}
% \label{icml-historical}
% \end{center}
% \vskip -0.2in
% \end{figure}

\begin{figure}[ht]
\vskip 0.2in
\begin{center}
\centerline{\includegraphics[width=0.8\linewidth]{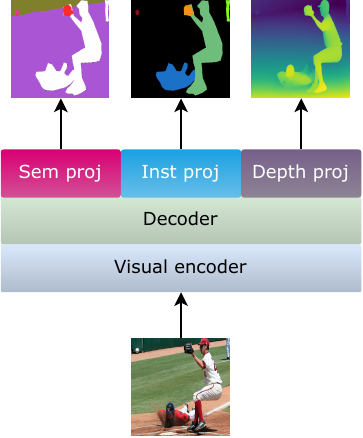}}
\caption{Overall model architecture in a multi-tasking setup.
The vision encoder and the shallow decoder are shared among all tasks.
The task specific projection heads are used only to map the decoded pixel embedding
to the task required dimension.
The depth prediction is an optional auxiliary task.}
\label{fig:arch}
\end{center}
\vskip -0.2in
\end{figure}

\begin{figure}[ht]
\vskip 0.2in
\begin{center}
\centerline{\includegraphics[width=0.8\linewidth]{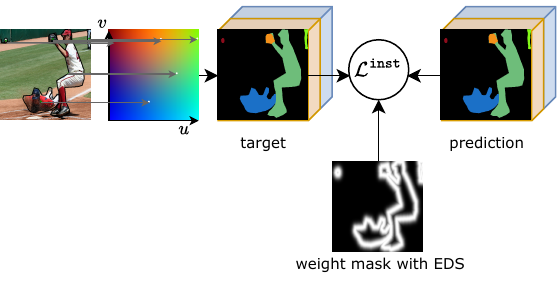}}
\caption{
Schematic explanation of how labeling instances work.
For each instance we find the $u$ and $v$ coordinates of its center of mass.
Based on these coordinates we label each pixel belonging to the instance.
The label is the concatenation of the positional embedding of the $u$ and $v$ coordinates, respectively.
Hence the two colored third dimension of the target and prediction.
Lastly, we apply a distance loss on this representation,
but only for pixels near object boundaries according to EDS (Section~\ref{sec:eds}).
}
\label{fig:centroid-encoding-chart}
\end{center}
\vskip -0.in
\end{figure}

\subsection{Architecture and task setup}
We present our work in a multi-tasking setup.
The panoptic segmentation task is split into two separate tasks -
semantic and class-agnostic instance segmentation.
The outputs of the two are combined to produce panoptic predictions.
% through majority voting.
We also add a third task to validate the generalist nature of our solution -
monocular depth prediction.
This task is optional, we report results with and without it.
% Additionally, in some of our experiments we add a third task -
% monocular depth prediction.
% With this we are able to showcase the generalist nature of our solution.

% The main focus in this work is panoptic segmentation.
% However, since our solution is generalist we work in a multi-tasking setup.
% With this choice we can solve panoptic segmentation by splitting it into two independent tasks -
% semantic and class-agnostic instance segmentation.
% On top of that, we add an additional task - monocular depth prediction.
% A post-processing procedure combines
% the predictions for semantic and instance segmentation tasks into panoptic output,
% through majority voting.

We use a standard deep encoder - shallow decoder architecture,
where both the encoder and the decoder are shared across the tasks.
The encoder backbone is DINOv2~\cite{dinov2},
while the decoder consists of four
transpose convolutional layers~\cite{long2015fully}.
The main purpose of the decoder
is to upscale the patch-level embeddings to pixel-level ones.
% to pixel-level ones - a necessity when dealing with dense predictions.
In essence the majority of the parameters are in the foundation vision backbone, which we fine-tune.
Finally, the output of the decoder is fed through a task specific projection layer,
in order to produce output in the correct format for each task.
See the overall architecture in Figure~\ref{fig:arch}.
Similar strategy,
without the multi-tasking setup has already been applied with DINO -
for example for semantic segmentation and depth~\cite{depthanything, depthanythingv2}.

Let the input image be $\mathbf{I} \in \mathbb{R}^{3\times H\times W}$.
The output of the encoder is $\mathbf{z} = f(\mathbf{I}) \in \mathbf{R}^{e\times H/p\times W/p}$,
where $p$ is the patch size of the encoder transformer~\cite{vit}
and $e$ is the embedding dimension throughout the encoder.
The output of the shared decoder is $\mathbf{h} = g(\mathbf{z}) \in \mathbb{R}^{d\times H\times W}$,
where $d$ is the embedding dimension used in the decoder.
Then the task-specific projection layer maps the $d$-dimensional embedding
into task specific output $\mathbf{\hat{y}}^\texttt{task}$.

%\subsection{Task-specific pixel-wise loss}
\subsection{Pixel-wise loss}
Using task-specific projection layers,
allows us to have a different number of output channels per task,
and thus have freedom with the choice of the loss.
With pixel-wise loss,
we can use any task-specific loss as long as it is computed on the level of the pixels.
The loss for each task $\mathcal{L}^t$ is simply average of the loss across all pixels,

\begin{equation}
    \mathcal{L}^t =
    \frac{\sum_{i=1}^{H}\sum_{j=1}^{W} w_{ij} L^t
    (\mathbf{y}_{ij}^t, \mathbf{\hat{y}}_{ij}^t)}
    {\sum_{i,j}w_{ij}},
\end{equation}
where $t \in \{ \texttt{sem}, \texttt{inst}, \texttt{depth} \}$,
$\mathbf{y}_{ij}$ is the $(i, j)$ pixel in $\mathbf{y}$.
$L^t(\mathbf{y}^t_{ij}, \mathbf{\hat{y}}^t_{ij})$
is the task loss for pixel $(i, j)$.
Note that we perform weighted average using the weight coefficients $w_{ij} \in [0,1]$.
This is how we perform Edge distance sampling EDS (Section~\ref{sec:eds}).
% For the semantic segmentation task we use standard cross-entropy loss.
% The loss for the class-agnostic instance segmentation based on positional embeddings
% is introduced in Section~\ref{sec:pos_embed}.
% Since we only use depth to show that our multi-tasking solution is generalist,
% we do not introduce any novel components or loss functions.
% See Appendix~\ref{appendix:depth_appendix} for more information on the loss function we use.

\subsection{Panoptic segmentation}
As already mentioned, we split panoptic segmentation into
semantic and class-agnostic instance segmentation.
% We split panoptic segmentation into two tasks -
% semantic segmentation and class-agnostic instance segmentation.
They are solved independently with averaged per-pixel loss.
Semantic segmentation is solved as a standard classification task
with a cross-entropy loss.
The output of the task specific head has $N$ channels,
where $N$ is the number of categories in the dataset.
The approach for instance segmentation is explained next.
% The approach for instance segmentation is through centroid regression
% using the positional embeddings of the centroid coordinates.

\subsubsection{Positional embeddings for centroid regression}
\label{sec:pos_embed}
We use the objects' centers of mass as a proxy for their instance segmentation labels.
% As mentioned earlier we tackle the class-agnostic instance segmentation task
% through encoding the instances via their centers of mass.
In other words,
for each pixel we predict the centroid coordinates for the instance it belongs to.
If a pixel does not belong to any labeled instance, we label it with a predefined \textit{void} encoding.
In this setup, the magnitude of the loss directly depends on the accuracy of the predicted centroid.
Note, that the centroids are only meant as proxies for clustering the pixels
into a set of instances.
% which belong to a single instance.
Beyond this, the accuracy of the predicted centroid is of no significance.
However, high accuracy in the centroid prediction is required
on the boundary between instances with close centers of mass.
With distance based loss,
the scale of the loss on the boundary between close-by instances
can be low because their respective centers of mass are close-by.
Even when the model is mistaken about which instance the pixel belongs to.
% With distance based loss, the scale of the loss on the boundary between close-by instances
% can be low even with poor separation between the two instances.
This can lead to imbalance during training.
A visual explanation of the source of this imbalance
can be seen in Figure~\ref{fig:distance-dependence}
accompanied with detailed explanation in Section~\ref{sec:motivation}.

To address this issue,
we propose to not compute the loss as a distance in the 2D space of the centroid coordinates.
Instead we predict the spectral positional embeddings of the centroid coordinates,
and compute the loss in this higher-dimensional space.
There the distances between the embeddings of the centroids are more balanced.
For example,
two centroids that are close in the image space can be far in the space of the
positional embeddings.
Importantly for us, the scale of the loss for mistakes in different parts of the image becomes more equal.
% To address the aforementioned issue,
% we propose to use the positional embedding of the centroid's coordinates,
% mapping them to a higher-dimensional space.
% This leads to more equalized distances between the encoding of centers of mass,
% independently of how close or far the centroids are in the image.
% In other words,
% the overall scale of the loss for mistakes in different parts of the image becomes more equal.

Concretely, let $u$ and $v$ be the coordinates of the center of mass of a given instance.
We first normalize them to be values in the range $[-1, 1]$.
Then we apply positional embeddings as in~\cite{vaswani2017attention,nerf}.
The final representation of each pixel is given by the concatenation of the embeddings for $u$ and $v$

\begin{equation}
    \mathbf{y}_{ij}^{\texttt{inst}} =
    \begin{cases}
    \texttt{concat}(\gamma(u), \gamma(v)), & (i, j) \in \mathbf{O}(u, v) \\
    \mathbf{0} \in \mathbb{R}^{4L} ,& (i, j) \text{ unlabeled},
    \end{cases}
\end{equation}

where pixel $(i, j)$ either belongs to object $\mathbf{O}(u, v)$ with centroid coordinates $(u, v)$
or is unlabeled.

The positional embedding $\gamma$ is defined as,

%\begin{equation}
%    \gamma(p) = \bigcup_{l=0}^{L-1} \left[\sin(2^l\pi p), \cos(2^l\pi p)\right] \in \mathbb{R}^{2L}
%\end{equation}

\begin{equation}
    \gamma(p) = \left(\sin(2^l\pi p), \cos(2^l\pi p)\right)_{l=0}^{L-1} \in \mathbb{R}^{2L}.
\end{equation}

It is a $2L$-dimensional vector,
where $L$ is the number of harmonics we use.
The output $\hat{\mathbf{y}}^\texttt{inst}$ for the instance segmentation task
is a $4L$-dimensional vector for each pixel.

The pixel loss for the instance segmentation task
is computed as the average euclidean distance
between the prediction and the target embeddings, of the $u$ and $v$ coordinates,
respectively.
It is

% \begin{equation}
%     {L}(\mathbf{y}_{ij}, \mathbf{\hat{y}}_{ij}) =
%     \frac{1}{2} \sum_{n=1}^2{
%     \Big\| (\mathbf{y}_{ij} - \hat{\mathbf{y}}_{ij})_{[2(\!n\!-\!1\!)L\!+\!1:2nL]} \Big\|_2}.    
% \end{equation}

\begin{equation}
\begin{split}
    {L}^{\texttt{inst}}(\mathbf{y}_{ij}, \mathbf{\hat{y}}_{ij}) =
    \frac{1}{2} (&\Big\|(\mathbf{y}_{ij} - \mathbf{\hat{y}}_{ij})_{[1:2L]}\Big\|_2 +\\
    &\Big\|(\mathbf{y}_{ij} - \mathbf{\hat{y}}_{ij})_{[2L+1:4L]}\Big\|_2).
\end{split}
\end{equation}

During inference,
the $u$ and $v$ coordinates are recovered from
$\hat{\mathbf{y}}_{ij}$ using the nearest neighbor search method,
between the prediction and positions in a discretized $uv$-grid space. 
This discretization is used to make post-processing simpler.

\subsubsection{Edge distance sampling}
\label{sec:eds}
Although we compute a pixel-wise loss,
we don't let all pixels contribute equally.
We only train on pixels close to object boundaries.
For each pixel $(i, j)$ we compute the euclidean distance $d_{ij}$ to the nearest boundary.
Then the weight of that pixel is given by

\begin{equation}
    w_{ij} = w_{min} + (1 - w_{min}) e^{-\frac{d_{ij}^2}{D^2}}.
\end{equation}

Here $w_{min}$ is the minimum weight for pixels that are far away from boundaries.
In our experiments $w_{min} = 0$.
The parameter $D$ determines how quickly the weight decreases when going away from a boundary.
We use a value of $D=20$ in our experiments.
Example weight masks are shown in Figures~\ref{fig:centroid-encoding-chart} and ~\ref{fig:expansion}.

Weight distribution in this manner
helps us to focus on the image regions with highest information.
It is akin to lateral inhibition in animal vision~\cite{livingstone2022vision,jernigan2018lateral}
which leads to higher activations only near edges.
Similar approach is used in other works~\cite{zhen2019learning,zhang2019deep,vsaric2023panoptic,zhu2019improving}
with the goal of focusing the training on the more challenging areas of the image.
Additionally, and more importantly to us, EDS addresses two key problems.
First, it mostly discards unlabeled regions when computing the loss for instance segmentation.
Second, it reduces the imbalance between small and large instances during training.
Both are explained in detail in Section~\ref{sec:motivation}.

\subsubsection{Additional losses.}
We use two additional losses,
total variation loss~\cite{rudin1992nonlinear} as a form of regularization
and generalized DICE loss~\cite{sudre2017generalised,crum2006generalized}.
The DICE loss is used only for instance segmentation.
Find more details about them in Appendix~\ref{appendix:losses}.

\subsection{Monocular depth prediction}
\label{sec:depth_loss}
We use depth to validate that our solution is generalist.
Since our encoder-decoder framework with pixel-wise prediction
is already suitable for depth,
we do not have to add any new components or loss functions.
Simply, we attach another projection layer to the shared decoder
and apply the standard loss functions following~\cite{depthanything}.
Similarly to DepthAnything,
we train in two stages - pre-training on relative depth with pseudo labels,
and fine-tuning on metric depth with ground truth labels.
The loss functions we use for depth
are explained in greater detail in Appendix~\ref{appendix:depth_appendix}.
The pseudo labels used during pre-training come from DepthAnythingV2.

% We use two different depth prediction tasks -
% metric depth on NYUv2~\cite{silberman2012indoor},
% and relative depth  using pseudo labels from DepthAnythingV2~\cite{depthanythingv2}.
% Relative depth is more suitable when pretraining with diverse scenes of different depth scales,
% for example indoor versus outdoor scenes.
% Metric depth is used to fine-tune on the specific benchmark.
% In our case this is NYUv2.
% With both tasks we use standard loss functions~
% \cite{birkl2023midas,depthanything,li2018megadepth}.
% More information can be found in Appendix~\ref{appendix:depth_appendix}.
% It is not a goal of this paper it introduce novel solutions for depth.
% We use the task to validate that our solution in generalist.

%% file: sec/4_rationale.tex
\section{Rationale}
\label{sec:motivation}
In this section
we explain the rationale behind our contributions,
positional embedding based loss for the class-agnostic instance segmentation
and Edge distance sampling (EDS).
Despite being simple,
they address important limitations of the proposed framework.

\begin{figure}[ht]
\vskip 0.2in
\begin{center}
\centerline{\includegraphics[width=0.8\linewidth]{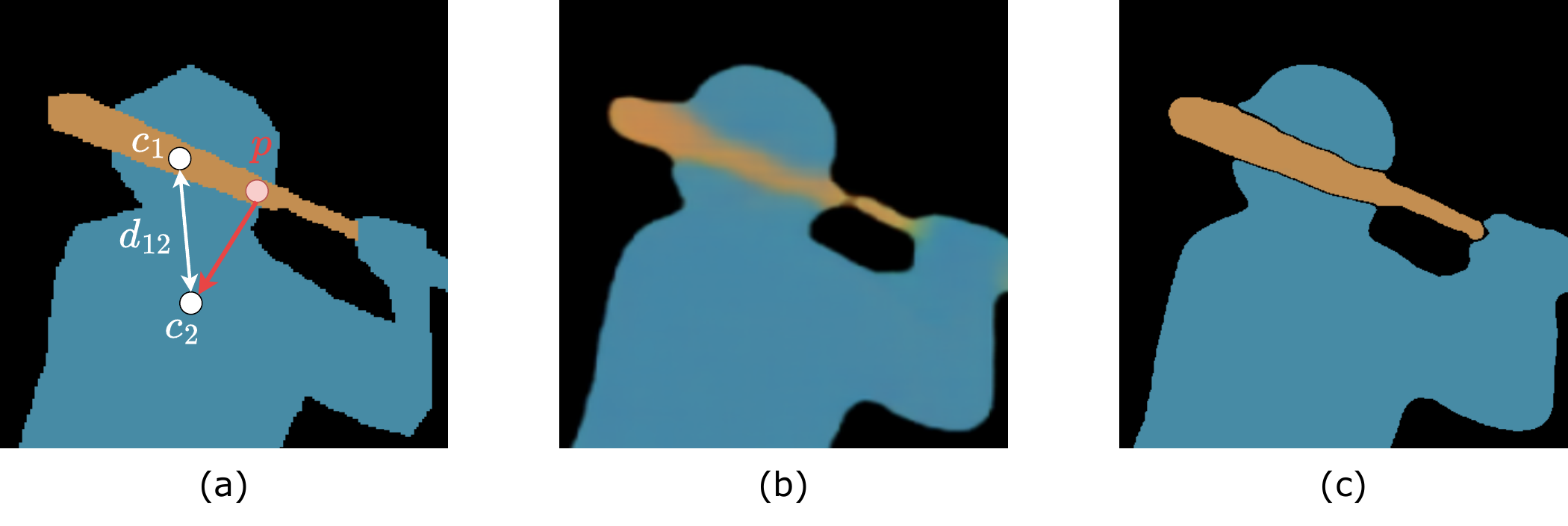}}
\caption{
With centroid encoding of the instances,
the scale of the loss for mistakes at the border between two instances
depends on how far apart the two centroids are.
Note, that here the distance $d_{12}$ is not very small for visualization purposes,
however, the model without PE based encoding (b) still struggles on the boundary.
You can also see the prediction from a model trained with PE based encoding (c).
}
 \label{fig:distance-dependence}
\end{center}
\vskip -0.2in
\end{figure}

\begin{figure}[ht]
\vskip 0.2in
\begin{center}
\centerline{\includegraphics[width=0.9\linewidth]{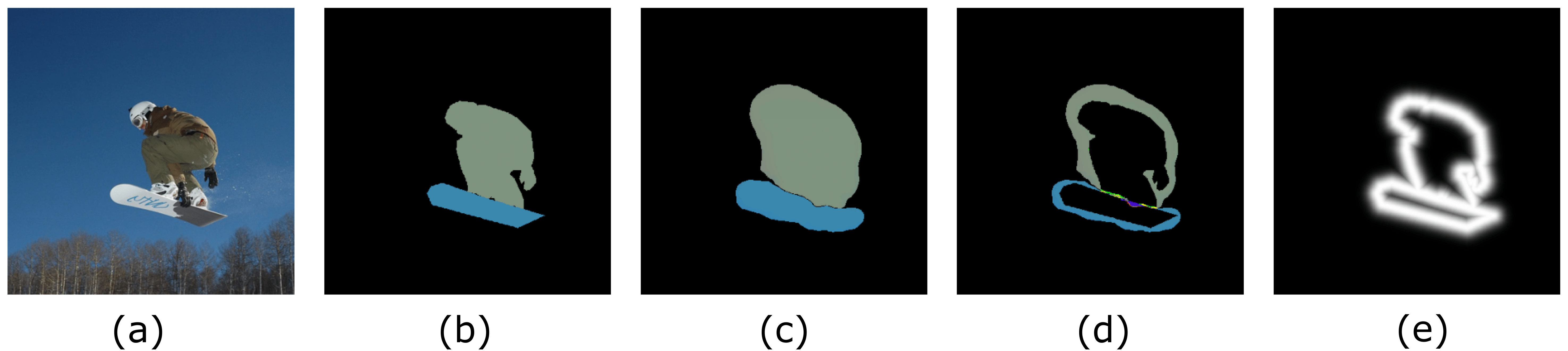}}
\caption{Instance expansion due to ignoring the void regions during training.
The prediction (c) is much expanded compared to the target (b).
The difference is shown in (d).
A solution for this issue is the proposed Edge Distance Sampling (EDS).
The weights obtained from EDS are shown in (e).}
\label{fig:expansion}
\end{center}
\vskip -0.2in
\end{figure}

\paragraph{Handling unlabeled regions.}
Instance segmentation datasets contain unlabeled or \textit{void} regions.
This is either due to pixels belonging to stuff categories,
or belonging to unlabeled objects.
Because of the latter, these void regions should be ignored when computing the loss.
However, with averaged pixel-wise loss,
if we ignore the unlabeled regions, we fail to teach the model where objects end.
This leads to a phenomena where the model's predictions tend to expand instances,
especially when they border an unlabeled region.
See an example in Figure~\ref{fig:expansion}.

Solutions like Painter~\cite{wang2023images}
treat the unlabeled \textit{void} regions as background.
This prevents the expansion issues,
but wrongly penalizes the model when it predicts instances that are not labeled.
When performing Edge Distance sampling (Section~\ref{sec:eds}) with $w_\text{min} = 0$,
we minimize the probability of such wrong penalization,
while successfully preventing the expansion problem.

\begin{figure}[ht]
\vskip 0.2in
\begin{center}
\centerline{\includegraphics[width=0.8\linewidth]{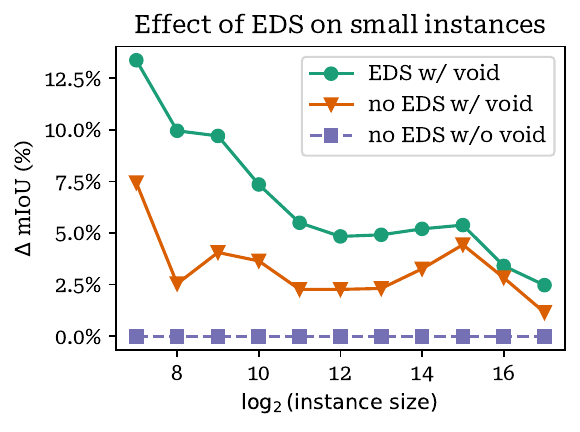}}
\caption{Effect of EDS on small instances.
This charts shows that EDS has disproportionately bigger effect on smaller instances.}
\label{fig:eds-small-impact}
\end{center}
\vskip -0.2in
\end{figure}

\paragraph{Imbalance due to instance sizes.}
Some of the most common problems when training with pixel-wise loss come from imbalance.
Imbalance can be caused by categories which are under-represented in the data.
But it can also be caused by some categories having instances of smaller size.
An example consequence of such imbalance is
when the model's loss is dominated by the small errors of already well predicted large objects,
instead of by the large errors of poorly predicted small objects.
This happens because large instances have much more pixels
and thus overwhelm the averaged loss.

Edge distance sampling (Section~\ref{sec:eds})
% successfully limits the effect of imbalance due to object size.
successfully reduces the imbalance due to object size.
Intuitively, this is easy to understand with a simple example.
Take two circles,
the first one with radius $R$, and the second one with radius $2R$.
% Let us say the first circle has radius $R$ and the second circle has radius $2R$.
The ratio of the areas of the two circles is $1:4$.
Hence, the second circle contributes 4 times more to the averaged loss.
However, edge distance sampling mostly samples around the circumference of the two circles.
Thus the contribution of the big circle will be roughly twice as large.
Our experiments confirm this intuition.
See the improvement caused by EDS plotted against object size in Figure~\ref{fig:eds-small-impact}.

\begin{figure}[ht]
\vskip 0.2in
\begin{center}
\centerline{\includegraphics[width=0.8\linewidth]{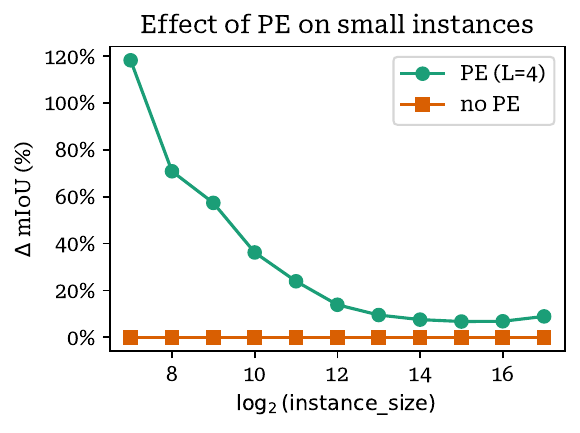}}
\caption{Effect of using PEs for the instance segmentation loss.
The improvement of the Intersection-over-Union is much stronger for smaller objects.}
\label{fig:pos-embeds-small-impact}
\end{center}
\vskip -0.2in
\end{figure}

\paragraph{Imbalance due to difference in the scale of the loss.}
Imbalance can also appear
if the scale of the loss is different at different parts of the images.
For the semantic segmentation task we use cross-entropy loss which is scale invariant.
However, for the class agnostic instance segmentation through centroid regression,
we use distance loss which is sensitive to the scale of the predicted values.
Let us look at this through the example in Figure~\ref{fig:distance-dependence}.
With EDS we are only computing the loss near object boundaries.
% With edge distance sampling we are only computing the loss near the boundary between two instances
% or the boundary between an instance and the unlabeled \textit{void} region.
In Figure~\ref{fig:distance-dependence}
point $p$ is near the boundary of the instances with centers $c_1$ and $c_2$.
Let us say that the model wrongly assigns point $p$ to the object with center $c_2$.
Even then, the scale of the loss will depend on the distance between $c_1$ and $c_2$.
% So even if we are mistaken to which of the two instances the point $p$ belongs to,
% the scale of the loss is determined by the distance between the instances' centroids.
It is interesting to note that small neighboring instances
are more likely to have their centroids close to each other.
So small instances which already contribute to the loss with fewer pixels,
now contribute even less because the scale of the loss is smaller than for large instances.

Using positional embeddings lifts the centroid coordinates to a high-dimensional space.
In a way, we are using the curse of dimensionality~\cite{bellman1959dynamic,bellman1961adaptive} in our favor.
In a high-dimensional space the distances between sets of points are likely to be larger,
leading to the contrast ratio of longest to shortest distance in a set of points being smaller.
For example, consider an $80\times80$ grid of $uv$-coordinates for the centroids.
% For example, consider the grid of $uv$-coordinates for the centroids in a $80\times80$ grid.
If we use L1 loss for direct regression of, say the $u$ coordinate,
then the ratio of the longest to shortest distance between any two points is approximately $80:1$.
The distance to the furthest point is 80 times larger than the distance to the nearest neighbor.
However, if we use PE based encoding with four harmonics ($L=4$),
then the contrast ratio will be roughly $5:1$.
This way the scale of the loss will be less dependent on how close the centroids are.

Another interesting observation here
is the fact that the largest boundary in the COCO dataset is between an instance and a \textit{void} region.
So it is very important how we encode the void region and how "far" it is to other instances.
If the encoding of the void region is far away from the instances,
this opens a potential imbalance where the boundaries between an instance and the void region dominate the loss.
We saw this empirically when training models without positional embeddings for the centroids.
Boundaries between instances, especially if they are small, were much harder to learn,
then boundaries between an instance and the \textit{void} region.
You can clearly see this in the second prediction of Figure~\ref{fig:distance-dependence}.

%% file: sec/5_experiments.tex
\section{Experiments}
\label{sec:experiments}

\newcommand{\cmark}{\ding{51}}%
\newcommand{\xmark}{\ding{55}}%

\begin{table}[t]
\caption{Panoptic segmentation results on COCO \texttt{val2017} split.
Our method achieves \textit{state-of-the-art} results among generalist methods,
while also performing well on depth.}
\label{tab:coco-results}
\vskip 0.15in
\begin{center}
\begin{small}
\begin{sc}
\begin{tabular}{r|c|c|c}
    \toprule
    Method & Backbone & Params & PQ $\uparrow$ \\
    \midrule
    \multicolumn{4}{l}{\textit{Specialized methods:\hfill}} \\
    MaskFormer& Swin-L & 212M & 52.7 \\
    % MaskFormer~\cite{cheng2021per} & Swin-L~\cite{swin} & 212M & 52.7 \\
    % Mask2Former~\cite{cheng2022masked} & Swin-L~\cite{swin} & 216M & $1024\times1024$ & 57.8 \\
    Mask2Former & Swin-L & 216M & 57.8 \\
    % Mask2Former~\cite{cheng2022masked} & Swin-L~\cite{swin} & 216M & 57.8 \\
    OneFormer & DiNAT-L & 223M &  58 \\
    % OneFormer~\cite{jain2023oneformer} & DiNAT-L~\cite{hassani2022dilated} & 223M &  58 \\
    % OneFormer~\cite{jain2023oneformer} & DiNAT-L~\cite{hassani2022dilated} & 223M & $1024\times1024$ & 58 \\
    kMax-DeepLab & ConvNext & 232M & 58.1 \\
    % kMax-DeepLab~\cite{yu2022kmax} & ConvNext~\cite{liu2022convnet} & 232M & 58.1 \\
    % kMax-DeepLab~\cite{yu2022kmax} & ConvNext~\cite{liu2022convnet} & 232M & $1281\times1281$ & 58.1 \\
    Mask-DINO & Swin-L & 223M & \textbf{59.4} \\
    % Mask-DINO~\cite{li2023mask} & Swin-L~\cite{swin} & 223M & \textbf{59.4} \\
    % Mask-DINO~\cite{li2023mask} & Swin-L~\cite{swin} & 223M & $1024\times1024$ & \textbf{59.4} \\
    \midrule
    \multicolumn{4}{l}{\textit{Generalist methods:\hfill}} \\
    UViM & ViT-L & 939M & 45.8 \\
    % UViM~\cite{kolesnikov2022uvim} & ViT-L~\cite{vit} & 939M & 45.8 \\
    % UViM~\cite{kolesnikov2022uvim} & ViT-L~\cite{vit} & 939M & $1280\times1280$ & 45.8 \\
    Painter & ViT-L & 303.5M & 41.3 \\
    % Painter~\cite{wang2023images} & ViT-L~\cite{vit,he2022masked} & 303.5M & 41.3 \\
    % Painter~\cite{wang2023images} & ViT-L~\cite{vit,he2022masked} & 303.5M & $448\times448$ & 41.3 \\
    Painter-MT & ViT-L & 303.5M & 43.4 \\
    % Painter~\cite{wang2023images} & ViT-L~\cite{vit,he2022masked} & 303.5M & 43.4 \\
    % Painter~\cite{wang2023images} & ViT-L~\cite{vit,he2022masked} & 303.5M & $448\times448$ & 43.3??? \\
    LDM-Seg & UNet & 851M & 43.3 \\
    % LDM-Seg~\cite{van2024simple} & UNet~\cite{ronneberger2015u} & 851M & 43.3 \\
    % LDM-Seg~\cite{van2024simple} & UNet~\cite{ronneberger2015u} & 851M & $512\times512$ & 43.3 \\
    Pix2Seq-$\mathcal{D}$ & ResNet & 94.5M & 50.3 \\
    % Pix2Seq-$\mathcal{D}$~\cite{chen2023generalist} & ResNet~\cite{resnet} & 94.5M & 50.3 \\
    % Pix2Seq-$\mathcal{D}$~\cite{chen2023generalist} & ResNet~\cite{resnet} & 94.5M & $1024\times1024$ & 50.3 \\
    Ours w/ depth & ViT-L & 304.3M & \textbf{54.5} \\
    Ours w/o depth & ViT-L & 304.3M & \textbf{55.1} \\
    % Ours & ViT-L~\cite{vit,dinov2} & 304.3M & $840\times840$ & \textbf{54.6??} \\
    % Result from brilliant-firecracker-2853 epoch 39
    \bottomrule
\end{tabular}
\end{sc}
\end{small}
\end{center}
\vskip -0.1in
\end{table}

\begin{table}[t]
\caption{Results for depth on NYU-v2.
The model used for this comparison is the same as used for COCO panoptic segmentation in
Table~\ref{tab:coco-results}.}
\label{tab:nyu}
\vskip 0.15in
\begin{center}
\begin{small}
\begin{sc}
\begin{tabular}{r|c|c|c}
    \toprule
    & \multicolumn{2}{c}{NYUv2} & COCO \\
    Method & RMSE $\downarrow$ & Abs. Rel. $\downarrow$ & PQ $\uparrow$ \\
    \midrule
    DINOv2-lin1 & 0.384 & - & \xmark \\
    DINOv2-lin4 & 0.333 & - & \xmark \\
    DINOv2-DPT & 0.293 & - & \xmark \\
    Painter & \textbf{0.288} & 0.080 & 43.3 \\
    Ours & 0.315 & \textbf{0.078} & \textbf{54.5} \\
    % Result from brilliant-firecracker-2853 epoch 39
    % Ours & 0.324 & 0.081 & \textbf{54.6} \\
    \bottomrule
\end{tabular}
\end{sc}
\end{small}
\end{center}
\vskip -0.1in
\end{table}

\paragraph{Datasets.}
We perform our experiments on three datasets.
Panoptic segmentation on MS-COCO~\cite{lin2014microsoft} and
Cityscapes~\cite{cordts2016cityscapes},
depth on NYUv2~\cite{silberman2012indoor}.
% We perform our experiments on three datasets - MS-COCO~\cite{lin2014microsoft,kirillov2019panoptic} for panoptic segmentation,
% Cityscapes~\cite{cordts2016cityscapes} for panoptic segmentation,
% and NYUv2~\cite{silberman2012indoor} for metric depth prediction.
MS-COCO consists of approximately 118K training and 5K validation images,
with 80 thing classes and 53 stuff classes.
Cityscapes consists of approximately 3K training and 500 validation high-resolution images.
It contains 19 classes of which 8 are thing classes.
For NYUv2 we use a 47K split for training and the official validation set consisting of 654 images.
They are labeled for metric depth from 0.001 to 10 meters.

\subsection{Setup and implementation details}

Although we work with dense prediction tasks,
we use output images of smaller size.
This is done for computational reasons.
As DINOv2 is trained with patch size of 14,
it outputs patch-level embeddings with resolution $\frac{H}{14}\times \frac{W}{14}$.
We upscale these 8 times, resulting in final output of size $8\frac{H}{14} \times 8\frac{W}{14}$.
For instance, if the input image is of size $840\times840$, then the output is of size $480\times480$.
This is similar to the approach used in Pix2Seq-$\mathcal{D}$~\cite{chen2023generalist}.

\paragraph{Implementation details.}
We use DINOv2~\cite{dinov2} as a vision encoder.
For the ablation studies we use the 12 layer ViT-Base model.
Our main results are reported using the 24 layer ViT-Large model.
The shared lightweight decoder consists of 3 transposed convolutional layers~\cite{long2015fully}
and one convolutional layer. Each of the first three layers upscales the representation $2\times$.
The input to the decoder comes from concatenating four output layers of the encoder.
When using the 24-layer ViT-Large encoder,
we take the output from layers 5, 12, 18 and 24.
When using the 12-layer ViT-Base encoder,
we take the output from layers 3, 6, 9 and 12.
We use AdamW~\cite{loshchilov2018decoupled} optimizer with learning rate of $5\times10^{-4}$.
The learning rate is decayed through a cosine schedule~\cite{loshchilov2016sgdr}
after a linear warm up.
We do not freeze the weight of the backbone,
but use layer decay to suppress the learning rate for the earlier layers.

\paragraph{Training setup.}
We train our models in three stages.
This is motivated by two reasons.
First, training a ViT-Large backbone with large images is computationally expensive.
That is why we first train, over many epochs,
with small image sizes and aggressive cropping augmentations.
Then we train for a small number of epochs with large image sizes with less or no cropping.
The second reason is that the class-agnostic instance segmentation task
is much more challenging than semantic segmentation.
Note, that the model needs to know the exact location of the center of mass of an instance
to predict it correctly.
That is why we first train only on the instance segmentation task and
then continue with the other tasks in a multi-tasking setup.
Additionally, when pretraining for depth,
we use a relative depth prediction task with pseudo labels from DepthAnythingV2~\cite{depthanythingv2}.

Here are the three stages concretely.
\underline{Stage 1}: Training is only on the class-agnostic instance segmentation task.
The input image size is $280\times280$.
Training runs for 400 epochs with effective batch size of $1024$.
% The learning rate is $5\times10^{-4}$ which is decayed using a cosine schedule~\cite{loshchilov2016sgdr}.
\underline{Stage 2}: Training is perform on all tasks. For semantic and instance segmentation we use MS-COCO.
Depth is optional in this stage.
When it is included, it is on the MS-COCO and NYUv2 datasets
with pseudo labels from DepthAnythingV2.
% For depth we use both MS-COCO and NYUv2 images with pseudo labels coming from DepthAnythingV2.
% We include COCO images here, because the NYUv2 dataset constist of image from only a few indoor scenes.
Training is done for 200 epochs with effective batch size of $512$.
The input image size is $336\times336$.
\underline{Stage 3:} This is the final finetuning stage with large images.
Training runs for 50 epochs with effective batch size of $16$, with images of size $1008\times1008$.
Since the Cityscapes dataset is much smaller we fine-tune on it separately for 400 epochs with images of size $784\times1568$.
The dataset is not included in the first two stages.

\paragraph{Inference and evaluation.}
For the depth prediction task we only need to scale the final output of each pixel.
For example, with NYUv2 we have a sigmoid after the last layer and thus the output is in the range $[0-1]$.
We have to multiply the output values by the maximum depth of the dataset (in this case $10$).
For panoptic segmentation we need to combine the semantic and instance segmentation predictions.
Since our model produces a continuous output for each pixel,
we have to perform a post-processing step.
It includes nearest-neighbor search to attach each pixel to one of a discrete set of possible centroids.
Then we perform non-maximal suppression similarly to Painter~\cite{wang2023images}.
The final step is to combine the class-agnostic instances with the semantic labels
using majority voting.
% It includes a simple non-maximal suppression algorithm for clustering the pixels into class-agnostic instances.
% Then we use majority voting from the semantic prediction to label each instance.

\subsection{Results}

% \begin{figure}[t]
%   \centering
%    \includegraphics[width=0.8\linewidth]{figures/eds-small-impact.pdf}
% 
%    \caption{Effect of EDS on small instances.
%    This charts shows that EDS has disproportionately bigger effect on smaller instances.}
%    \label{fig:eds-small-impact}
% \end{figure}
% 
% \begin{figure}[t]
%     \centering
%     \includegraphics[width=0.8\linewidth]{figures/pos-embeds-small-impact.pdf}
%     \caption{Effect of using PEs for the instance segmentation loss.}
%     It shows that PEs help the model improve for smaller instances more effectively.
%     \label{fig:pos-embeds-small-impact}
% \end{figure}

\paragraph{Panoptic segmentation on COCO.}
We report our results on the \texttt{val2017} split,
including a model trained only on panoptic segmentation on COCO,
and one trained in a multi-tasking setup with depth.
You can see the results in Table~\ref{tab:coco-results}.
The specialized methods we compare to are
MaskFormer~\cite{cheng2021per},
Mask2Former~\cite{cheng2022masked},
OneFormer~\cite{jain2023oneformer},
kMax-DeepLab~\cite{yu2022kmax} and
Mask-Dino~\cite{maskdino}.
While still not matching their performance,
we significantly shrink the gap between specialized and generalist solutions.

\begin{figure}[ht]
\vskip 0.2in
\begin{center}
\centering
\centerline{\includegraphics[width=0.8\linewidth]{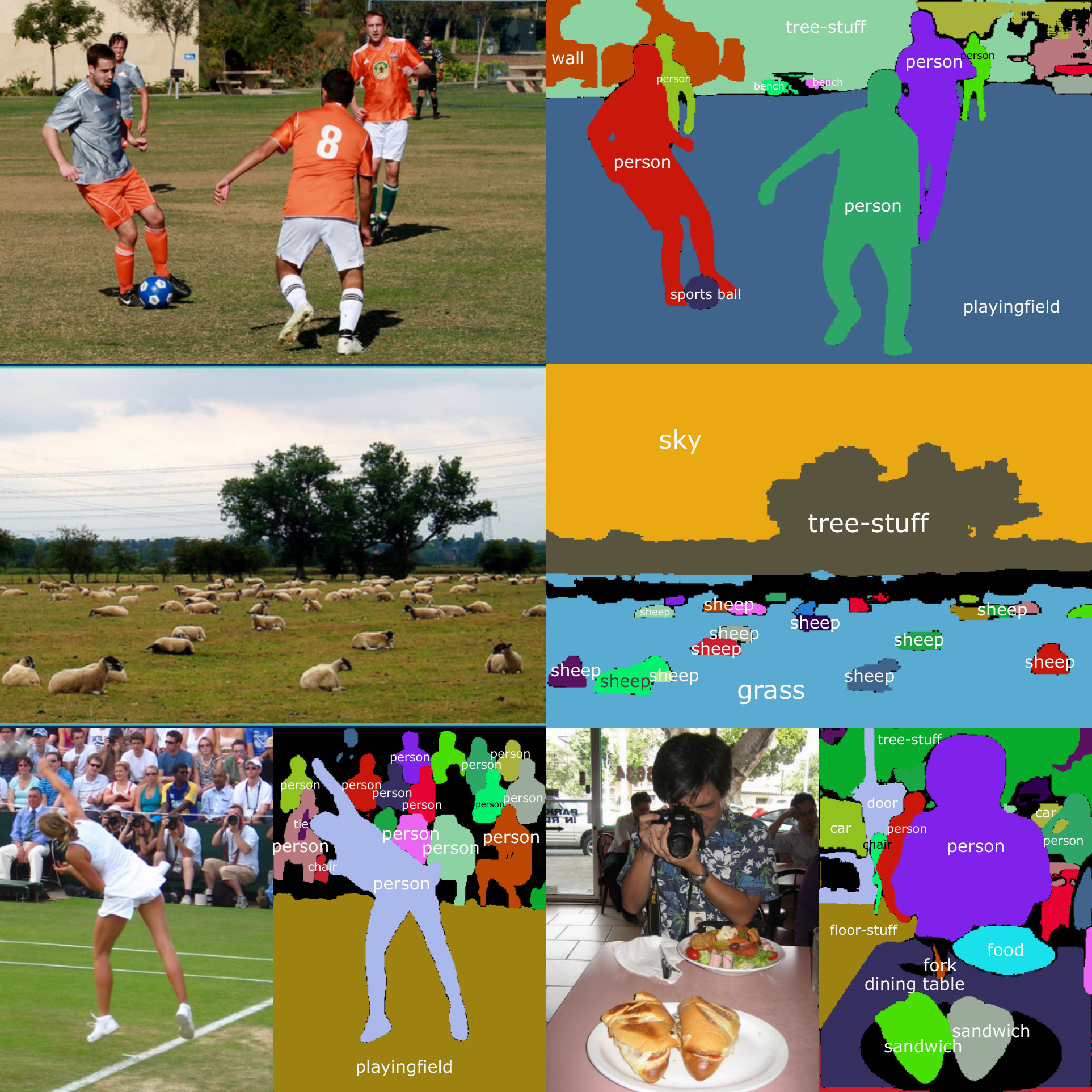}}
\caption{Qualitative example predictions for panoptic segmentation on the COCO dataset.}
\label{fig:qualy-preds}
\end{center}
\vskip -0.2in
\end{figure}

From the generalist approaches LDM-Seg~\cite{van2024simple}
and Pix2Seq-$\mathcal{D}$~\cite{chen2023generalist} are diffusion based.
Painter~\cite{wang2023images} is a multi-tasking model which performs centroid regression,
similarly to us.
See example predictions on the COCO dataset in Figure~\ref{fig:qualy-preds}.

% We report results on the COCO \texttt{val2017} split
% only in a multi-tasking setup that includes depth prediction.
% You can see comparison between our solution and \textit{state-of-the-art} methods in Table~\ref{tab:coco-results}.
% Our method achieves significantly better results on the panoptic segmentation task than other generalist methods,
% \textbf{54.5 PQ} against 50.3 for Pix2Seq-$\mathcal{D}$.
% The same model achieves strong depth performance on the NYUv2 dataset.
% We are getting close to bridging the gap to the specilized SOTA methods.
% We used images of size $840\times840$ for which the gridspace for centroid locations
% is $80\times80$.

\paragraph{Depth prediction on NYUv2.}
The results we report on NYUv2 are using the exact same model from the COCO panoptic segmentation results.
See its performance in Table~\ref{tab:nyu}.
We do not have depth specific contributions in our work.
The main goal of adding the depth task is to verify that our solution is generalist.
This is why we use finetuning DINOv2 as a baseline.
The performance of our model is on par,
while achieving very strong panoptic quality results.
DINO's linear head is most similar to our shared decoder,
however DINO achieve better results using a DPT head~\cite{dpthead} as decoder.
In our experiments the linear decoder performed better on the panoptic task,
thus it is our decoder of choice in all reported results.
% We could not make a model with a DPT-head decoder to work as strongly for panoptic segmentation.
% That's why we are leaving this opportunity for future work.
% See qualitative examples from the NYUv2 dataset in Figure~\ref{fig:depth-preds}.

\begin{table}[t]
\caption{Panoptic segmentation results on Cityscapes-val dataset.}
\label{tab:cityscapes}
\vskip 0.15in
\begin{center}
\begin{small}
\begin{sc}
\begin{tabular}{r|c}
    \toprule
    Method & PQ $\uparrow$ \\
    \midrule
    \multicolumn{2}{l}{\textit{Specialized methods:\hfill}} \\
    OneFormer~\cite{jain2023oneformer} & \textbf{68.5} \\
    % OneFormer~\cite{jain2023oneformer} & $512\times1024$ & \textbf{68.5} \\
    Mask2Former~\cite{cheng2022masked} & 66.6 \\
    % Mask2Former~\cite{cheng2022masked} & $1024\times2048$ & 66.6 \\
    kMax-DeepLab~\cite{yu2022kmax} & 68.4 \\
    % kMax-DeepLab~\cite{yu2022kmax} & $1025\times2049$ & 68.4 \\
    \midrule
    \multicolumn{2}{l}{\textit{Generalist methods:\hfill}} \\
    Pix2Seq-$\mathcal{D}$~\cite{chen2023generalist} & 64.0 \\
    % Pix2Seq-$\mathcal{D}$~\cite{chen2023generalist} & $1024\times2048$ & 64.0 \\
    Ours & \textbf{64.2} \\
    % Ours & $784\times1568$ & \textbf{64.2} \\
    \bottomrule
\end{tabular}
\end{sc}
\end{small}
\end{center}
\vskip -0.1in
\end{table}

\paragraph{Panoptic segmentation on Cityscapes.}
Our approach for Cityscapes~\cite{cordts2016cityscapes} is similar to the one used in Pix2Seq-$\mathcal{D}$~\cite{chen2023generalist}.
We take a \underline{Stage 2} model pretrained only on COCO and finetune for 400 epochs on Cityscapes.
This is done because the size of the dataset is too small.
See the results in Table~\ref{tab:cityscapes}.
As far as we are aware Pix2Seq-$\mathcal{D}$ is the only other generalist method
that reports results on Cityscapes.
We are able to improve on their performance
although marginally.
COCO is generally more challenging with 133 categories.
Cityscapes is less diverse (containing only driving scenes) and has only 19 classes.
It has high-resolution panoramic images.
We process images of size $784\times1568$.
When performing post-processing we label the instance of each pixel
based on nearest neighbor search from a $160\times320$ grid of possible centroid coordinates.
For comparison, the grid we use for COCO is $80\times80$.

% \begin{table}[t]
% \caption{Classification accuracies for naive Bayes and flexible
% Bayes on various data sets.}
% \label{sample-table}
% \vskip 0.15in
% \begin{center}
% \begin{small}
% \begin{sc}
% \begin{tabular}{lcccr}
% \toprule
% Data set & Naive & Flexible & Better? \\
% \midrule
% Breast    & 95.9$\pm$ 0.2& 96.7$\pm$ 0.2& $\surd$ \\
% Cleveland & 83.3$\pm$ 0.6& 80.0$\pm$ 0.6& $\times$\\
% Glass2    & 61.9$\pm$ 1.4& 83.8$\pm$ 0.7& $\surd$ \\
% Credit    & 74.8$\pm$ 0.5& 78.3$\pm$ 0.6&         \\
% Horse     & 73.3$\pm$ 0.9& 69.7$\pm$ 1.0& $\times$\\
% Meta      & 67.1$\pm$ 0.6& 76.5$\pm$ 0.5& $\surd$ \\
% Pima      & 75.1$\pm$ 0.6& 73.9$\pm$ 0.5&         \\
% Vehicle   & 44.9$\pm$ 0.6& 61.5$\pm$ 0.4& $\surd$ \\
% \bottomrule
% \end{tabular}
% \end{sc}
% \end{small}
% \end{center}
% \vskip -0.1in
% \end{table}

\begin{table}[t]
\caption{Results from the ablation studies for the effectiveness of PE based encoding
and EDS. You can see that both help independently,
but the best results are achieved when combining both.}
\label{tab:ablations}
\vskip 0.15in
\begin{center}
\begin{small}
\begin{sc}
\begin{tabular}{c|c|c}
    \toprule
    EDS & Centroid encoding & COCO PQ $\uparrow$ \\
    \midrule
    \xmark & Direct regression & \underline{38.01} \\
    \xmark & PE $L=4$ & 44.86 \\
    \cmark & Direct regression & 41.61 \\
    \cmark & Cross-entropy & 39.87 \\
    \cmark & PE $L=3$ & 47.38 \\
    \cmark & PE $L=4$ & \textbf{48.61} \\
    \cmark & PE $L=5$ & 48.12 \\
    \bottomrule
\end{tabular}
\end{sc}
\end{small}
\end{center}
\vskip -0.1in
\end{table}

\subsection{Ablations}
We perform an ablation study only for the panoptic segmentation task,
as we have no novel contributions for depth.
We evaluate the benefit of adding PE based loss and EDS
on the COCO dataset.
The results from the main study are shown in Table~\ref{tab:ablations}.
The models for the ablation study are trained using a smaller ViT-Base backbone with 12 layers.
The models are trained in a single stage of 200 epoch on smaller images of size $420\times420$.
This explains why the PQ metrics are lower than the ones reported in the experiments.
Despite that, it is clear that EDS and more importantly using PE based loss
is what enables the strong performance of our approach.
As a baseline we use a model without PE based centroid encoding,
i.e. direct regression, and without EDS.
You see that both help improve the performance,
and combining them leads to the biggest improvement.
More details on how we did direct regression without PEs
can be found the Appendix~\ref{appendix:alt_encodings}.

\paragraph{Addressing imbalance during training.}
Figures~\ref{fig:eds-small-impact} and~\ref{fig:pos-embeds-small-impact}
verify our intuition from Section~\ref{sec:motivation}
that EDS and PE based loss reduce the imbalance between small and large objects.
We plot the mean Intersection-over-Union for objects grouped by their area.
The y-axes show the relative improvement of models using the contributions against a baseline.
It is clear that EDS and especially the loss based on positional embeddings
bring disproportionately bigger improvement for smaller objects.
This supports our intuition about the imbalance during training.
The effect of the PE based loss for instance segmentation can be seen qualitatively
in Figure~\ref{fig:pos-embed-preds} of Appendix~\ref{appendix:qualitative}.
Additionally, Figure~\ref{fig:eds-small-impact} shows the impact of instance expansion
when excluding the unlabeled region during training.
The baseline excludes the void region when computing the loss for instance segmentation.
Including the void region when computing the loss brings improvement throughout all object sizes.
Adding EDS on top brings further improvements, especially for objects of smaller size.

% Figure~\ref{fig:eds-small-impact} and Figure~\ref{fig:pos-embeds-small-impact}
% show the effect of EDS and the PE based loss on the imbalance between small and large objects.
% This verifies our intuitions from Section~\ref{sec:motivation}.
% The plots show the improvement of mean Intersection-over-Union (mIOU) grouped by object size.
% Especially with the addition of PE based encoding for the centroids,
% we gain significantly more for smaller sized objects.
% This ties well to our observation that smaller instances
% are more likely to have their centers of mass nearby.
% Thus, not only do they contain fewer pixels,
% but the scale of the loss is likely to be smaller with direct regression.

%% file: sec/6_conclusion.tex
\section{Conclusion}
In this paper we presented a simple
but effective solution for panoptic segmentation.
We managed to shrink the gap between generalist and specialized methods,
by introducing a novel way to reduce the training imbalance
between small and large objects
when performing pixel-wise centroid regression.
We provided deep intuitive argumentation of our approach and backed it up with solid empirical evidence.

\paragraph{Limitations and future work.}
One should be careful when using the PE based loss.
Lifting to a very high-dimensional space
leads to better reduction of the imbalance,
however it makes training more challenging.
Making two high dimensional vectors to be close
gets significantly more difficult with the increase of dimensions.
We are interested to explore in the future
if the PE based loss can be beneficial for other tasks as well.

% We found that using PE based loss worked very well for the centroid regression task.
% This method should be used with care.
% Lift to a very high-dimensional space leads to better reduction
% of the imbalance during training.
% However, training is more challenging. A good balance must be found.
% It remains interesting for us whether this approach can be used for other tasks.

\section{Acknowledgements}
This research was partially funded by the dAIedge project (HORIZON-CL4-2022-HUMAN-02-02,
Grant Agreement Number: 101120726) and the Ministry of Education and Science of Bulgaria
(support for INSAIT, part of the Bulgarian National Roadmap for Research Infrastructure).
It was also supported with computational resources provided by Google Cloud Platform (GCP).

%% file: sec/a1_depth.tex
\section{Details for training depth models}
\label{appendix:depth_appendix}
For the depth prediction tasks we follow the approach used in DepthAnything v2~\cite{depthanythingv2}.
During pretraining in \underline{Stage 2}
we train on relative depth using pseudo labels coming from DepthAnythingv2~\cite{depthanythingv2}.
During finetuning in \underline{Stage 3}
we train on metric depth using NYUv2~\cite{silberman2012indoor} labels.
This allows us to use more diverse data while pretraining
and only learn metric depth for the indoor scenes of NYU at the very end.
The two tasks require different approach for the loss.
The loss functions we used are explained in the rest of this section.

\paragraph{Relative depth with pseudo labels.}
While pretraining we use images from both COCO~\cite{lin2014microsoft} and NYUv2~\cite{silberman2012indoor}.
Combined they contain a diverse set of outdoor and indoor scenes with varying scales.
This is why we follow the approach from MiDaS~\cite{birkl2023midas} and DepthAnything~\cite{depthanything,depthanythingv2},
where we work in disparity space.
Since both works do not share training code we had to reproduce the loss ourselves.
It will be shared with the rest of our source code.
In disparity space we work with $d$ which is $1/\text{depth}$ normalized to be between $0$ and $1$.
Then we use affine-invariant loss to compensate for the unknown scale of each image,

\begin{equation}
   \mathcal{L} = \frac{1}{HW}\sum_{i=1}^{HW}{\rho(d_i^* - d_i)},
\end{equation}

where $d_i^*$ and $d_i$ are the ground truth and predictions in the disparity space.
$\rho$ is the affine-invariant loss, i.e. $\rho(d_i^*, d_i) = |\hat{d}_i^* - \hat{d}_i|$.
$\hat{d}_i$ is the scaled and shifted version of $d_i$, i.e.

\begin{equation}
    \hat{d}_i = \frac{d_i - t(d)}{s(d)},
\end{equation}

where $t(d)$ and $s(d)$ are used to align the scale and median of the predictions and targets.

\begin{equation}
    t(d) = \text{median}(d), \texttt{ }
    s(d) = \frac{1}{HW}\sum_{i=1}^{HW}{|d_i - t(d)|}.
\end{equation}

\paragraph{Metric depth.}
Finally, in \underline{Stage 3},
we finetune the model on the metric labels of NYUv2.
We use a scale-invariant log loss similar to~\cite{eigen2014depth}.

\begin{equation}
    \mathcal{L} = \frac{1}{n}\sum_i{{d_i}^2} - \lambda
    (\frac{1}{n}\sum_i{d_i})^2,
\end{equation}

where $d_i = \log{y_i} - \log{y_i^*}$
is the log difference between the metric prediction and target.
We used a value of 0.5 for $\lambda$.

%% file: sec/a2_losses.tex
\section{Additional losses}
\label{appendix:losses}

As mentioned in the paper we use two additional losses to help us train a better model.

\paragraph{Total variation loss}
With edge distance sampling we don't include the inner parts of objects when computing the loss.
Also, the output of our model is continuous prediction for each pixel.
However, the desired output for both the semantic and class-agnostic instance segmentation tasks is discrete masks.
This is why we use total variation loss~\cite{rudin1992nonlinear} as a form of regularization.
This loss is penalizing the model when the output of a given pixel is different from its neighbors.
This is not desirable only for pixel that are at the border between different objects.
The total number of such pixels is negligible.
The definition of the total variation loss for a pixel $i, j$ is as follows

\begin{equation}
    l^{tv}_{i,j} = || \mathbf{y}_{i,j} - \mathbf{y}_{i-1,j} ||
    + || \mathbf{y}_{i, j} - \mathbf{y}_{i,j-1} ||
\end{equation}

Here $\mathbf{y}_{i,j}$ is the output of the model for pixel $i, j$.
The norm used depends on the main loss. If the main loss is $L_1$ or $L_2$, the norm here is the same.
For the instance segmentation task with positional embeddings the norm is the euclidean distance between the two embeddings.

When the main loss is cross entropy as in the semantic segmentation task,
the total variation loss is defined as

\begin{equation}
    \mathcal{L}_{tv} = \sum_{i=1}^{H-1} \sum_{j=1}^{W-1} \frac{1}{2} (\mathcal{H}(P_{i,j}, P_{i+1, j}) + \mathcal{H}(P_{i, j}, P_{i, j+1}))
\end{equation}

where $\mathcal{H}(a, b) = -a\log(b)$
and $P_{i, j}$ is the probability distribution defined by the logits of the output for pixel $(i, j)$.

\paragraph{DICE Loss}
With edge distance sampling we ignore most of the unlabeled pixels when computing the loss,
except in the globally applied total variation loss.
This is good, because we don't know what is the correct label there.
However, we know that whatever this label is it should not be the same as one of the already labeled instances.
To enforce this we use a generalized DICE loss\cite{sudre2017generalised,crum2006generalized}.
This loss is an approximation of the mean intersection over union.

\begin{equation}
    \mathcal{L}_{dice} = 1 - 2\frac{\sum_l w_l \sum_{i, j} r_{l,(i,j)} p_{l, (i, j)}}{\sum_l w_l \sum_{i,j} (r_{l, (i,j)} + p_{l, (i, j)})}
\end{equation}

Here, $l$ is iterating over all labeled instances,
while $i, j$ are iterating over all pixels.
$w_l$ is the weight for each labeled instance.
In our case $w_l = 1 / \text{Area}_{l}$.
$r_{l, (i, j)}$ is the participation of pixel $i, j$ to the instance $l$ according to the ground truth,
while $p_{l, (i, j)}$ is the prediction of the fuzzy participation of pixel $i, j$ to instance $l$.

\begin{equation}
    r_{l, (i, j)} = 
    \begin{cases}
    1 ,& \text{if pixel } i, j \text{ belong to instance } l \\
    0 ,& \text{otherwise}
    \end{cases}
\end{equation}

For the instance segmentation task with positional embeddings $p_{l, (i, j)}$ is a score between 0 and 1,
showing how likely the prediction for pixel $i, j$ belong to instance $l$.
Remember that $\bar{y}_{i,j} \in \mathbb{R}^{4L}$.
The first $2L$ dimensions are the positional embedding of the $x$ coordinate of the center of mass of the instance
to which the pixel belongs.
While the last $2L$ dimensions are the positional embedding of the $y$ coordinate.
Let $d_{i,j}^x$ and $d_{i,j}^y$ be the euclidean distance between the ground truth and prediction
for the embeddings of the $x$ and $y$ coordinates, respectively.
Then,

\begin{equation}
    p_{l,(i,j)} =  e^{-({d_{i,j}^x})^2} e^{-({d_{i,j}^y})^2}
\end{equation}

%% file: sec/a3_encodings.tex
\section{Alternative encodings for instance segmentation}
\label{appendix:alt_encodings}

Here we explain the alternatives we used to positional embeddings when regressing to the centroid
for encoding the instances.
There are two options, regression to the $u, v$ coordinates of the centroid and classification with cross entropy loss.

\paragraph{RGB encodings for centroid regression.}
Painter~\cite{wang2023images} used RGB encoding instead of directly predicting the coordinates of the centroid.
He have a similar approach in out ablation study.
We can directly regress to the $u$ and $v$ coordinates of the centroid,
but then it is not clear how to encode the void pixels.
Instead we use a 3-dimensional RGB encoding,
where the red channel was equal the $u$ coordinate of the centroid normalized to be in the region $[0, 1]$,
the green channel was similarly equal to the $v$ coordinate,
and the blue channel was $1$ only when the pixel was part of an instance.
The encoding for the \textit{void} pixels is pure black $(0, 0, 0)$.
This is the method referred to as \textbf{Direct centroid regression} in
Table~\ref{tab:ablations}.

\paragraph{Cross entropy}
We divide the image in $80\times80$ grid cells and consider all centroids within a given cell to be the same.
This allows us to solve the class-agnostic instance segmentation as a classification problem with 6400 categories.
Performing this was too computationally expensive as we need 6400 logits for each pixel.
Instead we classified independently the $u$ and $v$ coordinates of the center of mass
into 80 classes.
This is a much more computationally efficient solution.
Performance is not good,
because the cross entropy loss has no concept of distance.
Thus small mistakes when the center of mass is challenging to compute,
or when the ground truth mask is not precise,
lead to problems with training.

%% file: sec/a4_qualitative.tex
\section{Qualitative examples}
\label{appendix:qualitative}
\begin{figure*}
    \centering
    \includegraphics[width=0.99\linewidth]{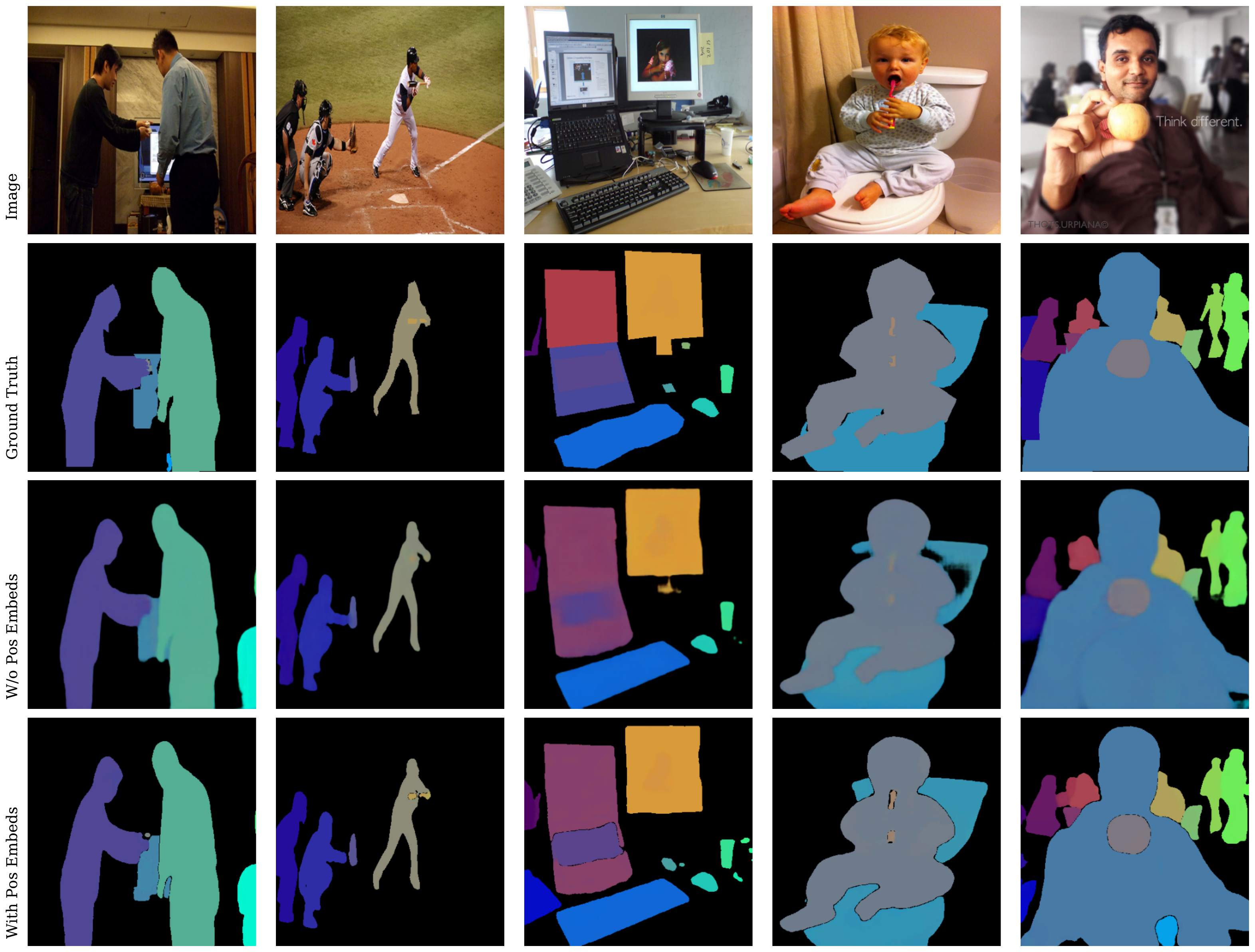}
    \caption{Comparing instance segmentation predictions with and without positional embeddings.
    It is clear how the boundaries between instances are much less blurry with positional embeddings.
    When training without positional embeddings the model learn the boundaries with void the best,
    but struggle with boundaries between objects, especially if the centers of mass are close.
    Both of these models are trained with DinoV2-Base backbone with image size $420\times420$.
    }
    \label{fig:pos-embed-preds}
\end{figure*}

Here we saw visual demonstration of the effect of positional embedding based loss for instance segmentation.
Figure~\ref{fig:pos-embed-preds} shows examples predictions from the models
used for the ablation studies.
You can clearly see that adding PE based encoding leads to much better defined boundaries.

